# Image Captioning in News Report Scenario

Tianrui Liu[1, *], Qi Cai[2], Changxin Xu[3], Bo Hong[4], Jize Xiong[5], Yuxin Qiao[6], Tsungwei Yang[7]

[1] Department of Electrical and Computer Engineering, University of California San Diego, USA
[2] Computer Science and Engineering, University of North Texas, Denton, USA
[3] Computer Information Technology, Northern Arizona University, Flagstaff, USA
[4] Computer Information Technology, Northern Arizona University, Flagstaff, USA
[5] Computer Information Technology, Northern Arizona University, Flagstaff, USA
[6] Computer Information Technology, Northern Arizona University, Flagstaff, USA
[7] Computer Science, Tunghai University, Taichung, Taiwan

[*] Corresponding author: Tianrui Liu (Email: tianrui.liu.ml@gmail.com )

**Abstract:** Image captioning[1] strives to generate pertinent captions for specified images[56], situating itself at the crossroads of Computer Vision (CV) [2] and Natural Language Processing (NLP)[54]. This endeavor is of paramount importance with far-reaching applications in recommendation systems, news outlets, social media, and beyond. Particularly within the realm of news reporting, captions are expected to encompass detailed information, such as the identities of celebrities captured in the images. However, much of the existing body of work primarily centers around understanding scenes and actions.[3] In this paper, we explore the realm of image captioning specifically tailored for celebrity photographs, illustrating its broad potential for enhancing news industry practices. This exploration aims to augment automated news content generation, thereby facilitating a more nuanced dissemination of information.[57] Our endeavor shows a broader horizon, enriching the narrative in news reporting through a more intuitive image captioning framework.

## 1. Introduction

Image captioning is a typical topic that bridges computer vision[6] and natural language processing.[10,42] In image captioning task, we aim to generate relevant captions for given images. As a very application-oriented task, there has been a wealth of research on image captioning.[12] This technique has been widely used in many fields: traffic detection[13,24], material analysis[22], communications deployment[35,39], aerial search[37], language model designing[38], embedding development[40,41]. Meanwhile, there are much less research that considers generating captions for specific names. We believe this problem is important in the settings of news report. In such scenarios, captions generated by algorithms should take the faces into account.[4] say we intend to generate descriptions like "Obama is delivering a speech".[36] Common method may only be able to generate sentences like "a man is delivering a speech" and is unacceptable apparently.[25]

In this project, we shed light on the problem of image captioning in scenarios where celebrities appear[50] and propose a combined method to solve this problem.[29] Our algorithms take three step to generate the ultimate sentences: (1) For the given image (with celebrities appear), we employ a common encoder-decoder architecture for image captioning and generate captions without names.[30] (2) We use MCTNN and Resnet[34] to get the names of faces appearing in the image. (3) We parse the output sentence in step(1) and replace the parts with celebrity names accordingly. Extensive experiments show the feasibility of our method in many simple scenarios.

The structure of the report are as follows: Section 2 defines the subtask of our project, which are comprised of Image captioning, Face recognition, Noun phrase(NP) chunks matching.[51] Section 3 discusses the details of our methods and implementations. Our experiment and results are shown in Section 4. In section 5, we conclude our approach and further discuss the strength and weakness of our methods.

## 2. Problem Definition

### 2.1. Image Captioning

The goal of image captioning is to generate relevant description for given images. The problem can be thought two-fold as it connects computer vision and natural language processing: 1) use an encoder architecture (CNN, Transformer)[32] to process the image; 2) use a decoder to decode the encoded image representation into sentences.[16] The model is trained in a supervised learning pattern[11] with some existing image to caption (ground truth) pairs. In the training stage we maximize the similarity of generated caption with ground truth,[14] and in the testing stage we decode the encoded image representation directly to obtain the outcome.

In recent years, various efforts have been made in the image captioning tasks and the performance of caption generation has improved significantly thanks to more efficient encoder-decoder architectures.[20]

### 2.2. Face recognition in images

Facial recognition is the task of making a positive identification of a face in a photo against a pre-existing database of faces. It begins with detection - distinguishing human faces from other objects in the image - and then works on identification of those detected faces, which can be seen as a classification problem. [15]

In general, face recognition is a well-defined and relatively mature field. There has been many theoretical analysis[21] and successful engineering practice that obtain very high accuracy. In this task, we employ MTCNN to obtain the bounding box for faces and use a Resnet pretrained with vggface2 to complete the classification tasks.[43]

### 2.3. Celebrity-aware image captioning

Celebrity-aware image captioning problem combines both image captioning and face recognition.

## 3. Approach

Our pipeline has three parts:

1. Image captioning: Image captioning modules do supervised training on datasets[49] (we use Flickr 8k/3k) and generate common (no specific names) captions in the testing stage.[7]

2. Face recognition: Face recognition employs MTCNN network to obtain bounding boxes for all appeared faces, and do classification using Inception\_v1 pretrained on vggface2.[31]

3. Noun phrase matching: Noun phrase matching module uses NLP packages and some rules to obtain noun phrase chunks (also needs to be people-related) like "a man" or "a young asian boy" and replaces them with the names derived in phrase II and generates the final outputs.[26]

The overall architecture of our pipeline is shown in figure 1.

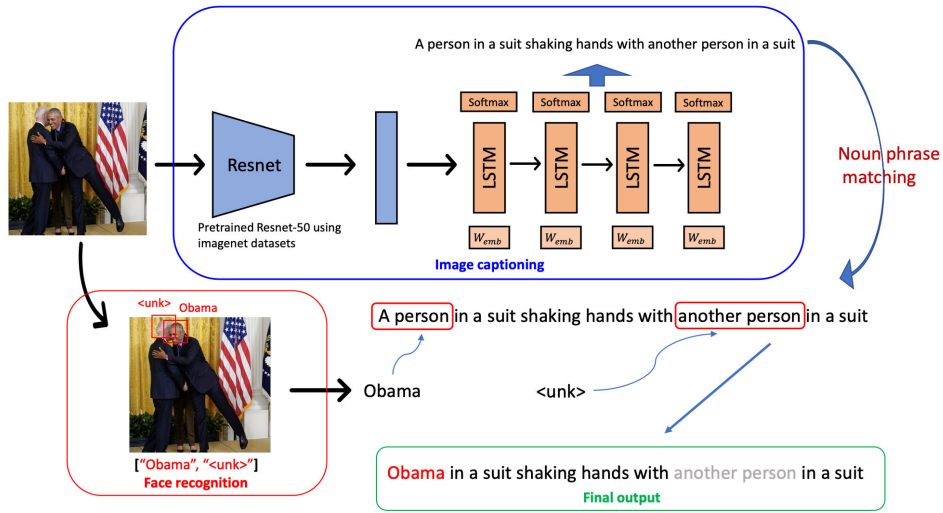

**Figure 1.** Overall image captioning architecture

### 3.1. Image captioning

In the image captioning step, we employ the widely-used encoder-decoder image captioning architecture. For the image side, we use a pretrained Resnet-50[53] without the last pooling and linear layer as encoder[52] and obtain a Tensor with shape [batch_size, 49, 2048]. [8]For decoder, we use a LSTM. To improve the performance of encoder-decoder, we add an Bahdanau attention layers[9] to calculate the attention between encoder outputs and initial states of the decoder.[5]

The model is trained in a supervised learning setting: we first trained the parameters of LSTM to minimize the loss between generated sentences and the ground true captions. In the testing stage, we generate captions using the optimal parameters and evaluate the results by human. The architecture of the image captioning module is shown in figure 2.[27]

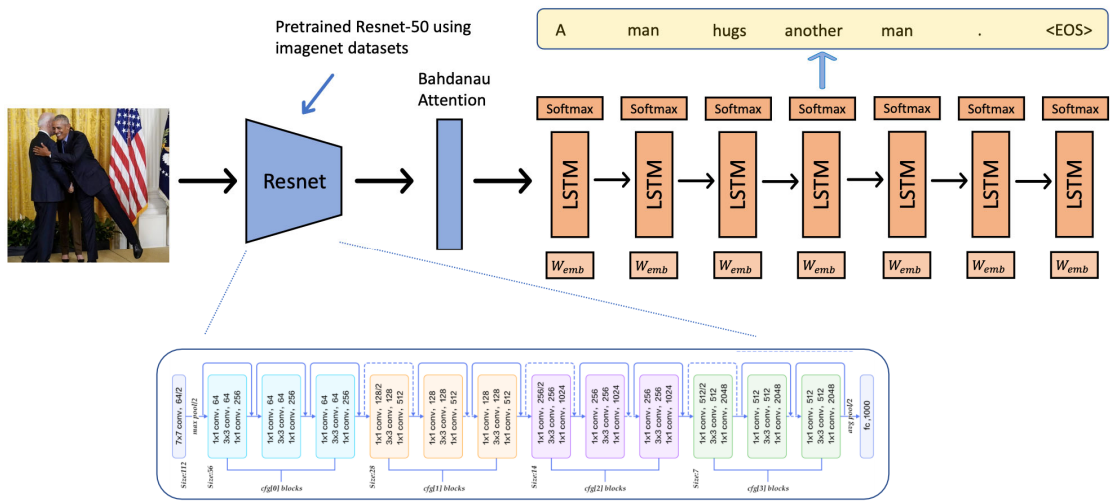

**Figure 2.** Image captioning module

## 3.2. Face recognition

Since our input images are scenes of behaviors instead of some simple human faces, we need to first extract face regions before performing face classification tasks. We follow the work of MTCNN [10] to complete this task: we first employ the MTCNN architecture to draw faces' bounding boxes and then use an Inception_Resnet_v1 to obtain the name classes. The architecture of our face recognition networks are shown in picture 3.[28]

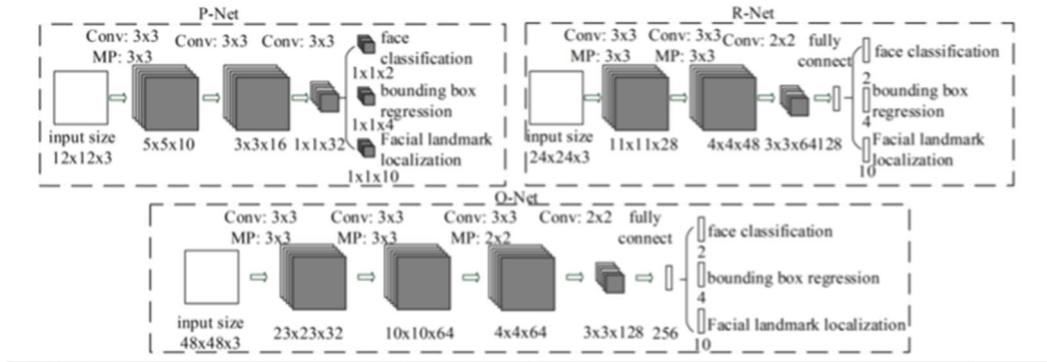

**Figure 3.** MTCNN architecture

Simply speaking, MTCNN are three cascaded convolutional networks: a Proposal Network (P-Net), a Refinement Network (R-Net) and an Output Network (O-Net). Firstly, candidate windows are produced through a fast P-Net. After that, we refine these candidates in the next stage through a R-Net. In the final stage, the O-Net produces final bounding box and facial landmarks position.

## 3.3. NP chunk matching

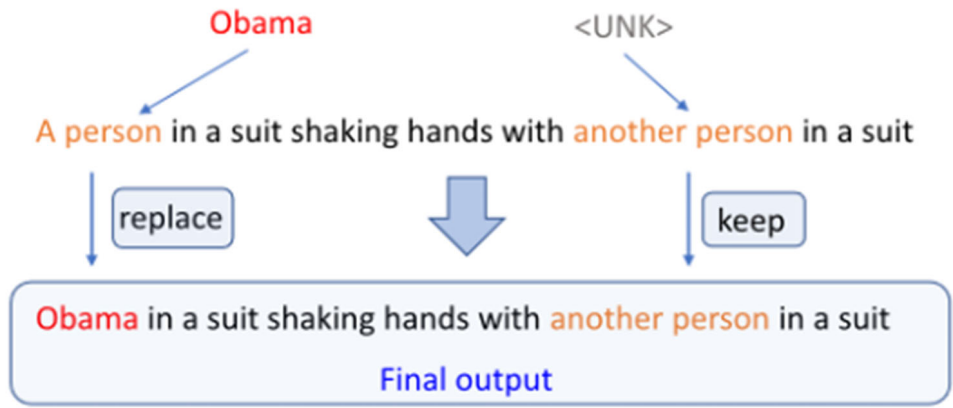

**Figure 4.** NP chunk matching

The last step is the alignment between celebrity names and the NP chunk in the generated captions. We parse the captions with NLTK and Spacy packages and obtain the people-related NP chunks. In general (sequence exchangeable) cases, sequence of the chunks are unimportant[55] so we simply sequentially align names with chunks. Using regular expression, we replace NP chunks like "A man", "The woman", "Two young boys" with celebrity names we derived in the face recognition step. Some corner cases need to be carefully coped with in this step, and in some cases it may be hard to match (or need more complicated rules).[58] For detailed implementation, please see refer to our code.

In nonexchangeable cases, problems are much more challenging. We intend to align the names and faces by considering the position relation between image2text attention matrix and recognized celebrity faces. Ideally, the pixel related to faces will have large attention weight, whose position will be helpful for the alignment. However, we find that the sequence of NP chunks in the datasets[59] are usually not in accordance with the position of faces. What's more, the attention we use in out experiment are not precise enough.[48] Due to those limitations, our method is limited to solving exchangeable cases, unless we alter the architecture.

## 4. Experiment

We use the following datasets in our project (due to stricter policy on privacy launched in recent years, we do not have access to larger datasets.

Flickr 8k/30k: contains about 8,000 images collected from Flickr, together with 5 reference sentences for each image provided by human annotators.

COCO Captions: contains over one and a half million captions describing over 330,000 images. For the training and validation images, five independent human generated captions are be provided for each image.

We carry out supervised learning using Flickr 8k/30k or Vggface2 for captioning. Overally speaking, our pipeline can obtain a very good results with an accuracy performance over

90%, which demonstrates the power of combining the deep neural networks and our face recognition modules.[47]

## 5. Conclusion and Discussion

### 5.1. Discussion

While our pipeline can solve the captioning problems in many cases, there are some cases we fails. The limitations of our method are as follows:

Mediocre generation performance: The generation performance is not exceedingly good because we use a rather small encoder decoder architecture in our model. In addition, we employs 30,000 images together with 150,000 captions in the training process.[44] Comparing to the massive datasets used in state-of-the-art language and computer vision models, this is still a rather standard number. As our image and caption data directly decides the volume of the vocabulary dictionary, different sentences patterns and object types, the poor performance obtained by limited datasets is within our expectation. In the future, we may attempt to increase the dataset size[19] and use more powerful encoder-decoder architectures (like using pretrained Bert for the captioning part)[17]

Inaccurate NP chunk matching: As we mentioned in section 4, at present our method are not capable to deal with immutable type celebrity image captioning problems due to the resolution of attention we uses and the inaccurate word sequence of Flickr datasets.[45] And in some cases, we fail to perform the matching.

There are potential solutions.

1. Use more sophisticated multi-modality approach. Particularly there are much research like CLIP that proposes to connect texts with images.

2. Improve the quality of the datasets. Some datasets contains anchor boxes for each object and person in the image, which can be utilized for a more precise matching. Also, we need to carefully look into the sequence of vocabularies in the sentence.

3. Consider the entire task jointly. If we can obtain image captioning datasets containing the celebrities' name, we can customize certain loss and improve the model's ability to recognize faces. Joint model may have better grammatical accuracy then our separate-steps pipeline.[18,46]

### 5.2. Conclusion

In this paper, we use a combined pipeline to conduct image captioning for celebrities. Our architecture uses a CNN and RNN based encoder-decoder to process the input images and generate captions. At the same time, MTCNN networks draws the bounding boxes for faces and Inception network performs classification task. Lastly, we employ NLP packages like NLTK and some engineered rules to implement NP chunk matching. [23]

The incorporation of such a pipeline can significantly abbreviate the time-to-market, while ensuring a high standard of accuracy and relevance in generated content.[33] Our endeavors lay a solid groundwork, beckoning a new era of intelligent, automated news generation systems attuned to the dynamic demands of the modern digital media.